\documentclass[11pt,a4paper]{article}
\usepackage[hyperref]{emnlp-ijcnlp-2019}
\usepackage{times}
\usepackage{latexsym}
\usepackage{multirow}
\usepackage{booktabs}
\usepackage{graphicx}
\usepackage{amssymb}
\usepackage{cleveref}
\usepackage{tikz}
\usetikzlibrary{tikzmark, positioning, fit}
\usetikzlibrary{decorations.pathreplacing, calc}

\usepackage{url}
\usepackage{linguex}
\alignSubExtrue 
\aclfinalcopy 

\def\newterm#1{\textit{#1}}

\crefformat{section}{\S#2#1#3}
\crefformat{subsection}{\S#2#1#3}

\title{Investigating BERT's Knowledge of Language:\\ Five Analysis Methods with NPIs}

\author{
 Alex Warstadt,$^{\dagger,1,2}$
 Yu Cao,$^{\dagger,3}$
 Ioana Grosu,$^{\dagger,2}$
 Wei Peng,$^{\dagger,3}$
 Hagen Blix,$^{\dagger,1}$
 Yining Nie,$^{\dagger,1,2}$\\\bf
 Anna Alsop,$^{\dagger,2}$
 Shikha Bordia,$^{\dagger,3}$
 Haokun Liu,$^{\dagger,3}$
 Alicia Parrish,$^{\dagger,2,3}$
 Sheng-Fu Wang,$^{\dagger,3}$
 Jason Phang,$^{\dagger,1,3}$\\\bf
 Anhad Mohananey,$^{\dagger,1,3}$
 Phu Mon Htut,$^{\dagger,3}$
 Paloma Jereti{\v c},$^{\dagger,1,2}$
 and Samuel R. Bowman\\
 New York University\AND\rm
 {\footnotesize $^\dagger$Equal contribution with roles given below; order assigned randomly. Correspondence: \href{mailto:bowman@nyu.edu}{\tt bowman@nyu.edu}} \\
 {\footnotesize
 $^1$Framing and organizing the paper~~$^2$Creating diagnostic data~~$^3$Constructing and running experiments
 }}

\date{}

\begin{document}
\setlength{\SubExleftmargin}{-1em}
\maketitle
\begin{abstract}
Though state-of-the-art sentence representation models can perform tasks requiring significant knowledge of grammar, it is an open question how best to evaluate their grammatical knowledge.
We explore five experimental methods inspired by prior work evaluating pretrained sentence representation models. We use a single linguistic phenomenon, negative polarity item (NPI) licensing in English, as a case study for our experiments. NPIs like \emph{any} are grammatical only if they appear in a \textit{licensing environment} like negation (\emph{Sue doesn't have any cats vs. *Sue has any cats}). This phenomenon is challenging because of the variety of NPI licensing environments that exist. 
We introduce an artificially generated dataset that manipulates key features of NPI licensing for the experiments. We find that BERT has significant knowledge of these features, but its success varies widely across different experimental methods. We conclude that a variety of methods is necessary to reveal all relevant aspects of a model's grammatical knowledge in a given domain.



\end{abstract}

\section{Introduction}

Recent sentence representation models have attained state-of-the-art results on language understanding tasks, but standard methodology for evaluating their knowledge of grammar has been slower to emerge. Recent work evaluating grammatical knowledge of sentence encoders like BERT \citep{devlin2018bert} has employed a variety of methods. For example, \citet{shi2016does}, \citet{ettinger2016probing}, and \citet{tenney2018what} use probing tasks to target a model's knowledge of particular grammatical features. \citet{marvin2018targeted} and \citet{wilcox-2019} compare language models' probabilities for pairs of minimally different sentences differing in grammatical acceptability. \citet{linzen2016assessing}, \citet{warstadt-2018}, and \citet{kann2019verb} use Boolean acceptability judgments inspired by methodologies in generative linguistics. However, we have not yet seen any substantial direct comparison between these methods, and it is not yet clear whether they tend to yield similar conclusions about what a given model knows.

We aim to better understand the trade-offs in task choice by comparing different methods inspired by previous work to evaluate sentence understanding models in a single empirical domain. We choose as our case study negative polarity item (\newterm{NPI}) licensing in English, an empirically rich phenomenon widely discussed in the theoretical linguistics literature \citep[a.o.]{kadmon-landman1993,giannakidou1998,chierchia2013}. NPIs are words or expressions that can only appear in environments that are, in some sense, \textit{negative} \citep{fauconnier1975,ladusaw1979,linebarger1980grammar}. For example, \textit{any} is an NPI because it is acceptable in negative sentences \ref{gram} but not positive sentences \ref{ungram}; negation thus serves as an NPI \newterm{licensor}. NPIs furthermore cannot be outside the syntactic \newterm{scope} of a licensor \ref{nosc}. Intuitively, a licensor's scope is the syntactic domain in which an NPI is licensed, and it varies from licensor to licensor. A sentence with an NPI present is only acceptable in cases where (i) there is a licensor---as in \ref{gram} but not \ref{ungram}---and (ii) the NPI is within the scope of that licensor---as in \ref{gram} but not \ref{nosc}.

\ex. \label{gram} 
Mary has\textbf{n't} eaten \textit{any} cookies.

\ex. \label{ungram}
*Mary has eaten \textit{any} cookies.

\ex. \label{nosc}
*\textit{Any} cookies have\textbf{n't} been eaten.

We compare five experimental methods to test BERT's knowledge of NPI licensing. We consider: (i) a \newterm{Boolean acceptability classification} task to test BERT's knowledge of sentences in isolation, (ii) an \newterm{absolute minimal pair} task evaluating whether the absolute Boolean outputs of acceptability classifiers distinguish between pairs of minimally different sentences that differ in acceptability and each isolate a single key property of NPI licensing, (iii) a \newterm{gradient minimal pair} task evaluating whether the gradient outputs of acceptability classifiers distinguish between minimal pairs, (iv) a \newterm{cloze test} evaluating the grammatical preferences of BERT's masked language modeling head, and (v) a \newterm{probing task} directly evaluating BERT's representations for knowledge of specific grammatical features relevant to NPI licensing.

We find that BERT does have knowledge of all the key features necessary to predict the acceptability of NPI sentences in our experiments. However, our five methods give meaningfully different results. While the gradient minimal pair experiment and, to a lesser extent, the acceptability classification and cloze tests indicate that BERT has systematic knowledge of all NPI licensing environments and relevant grammatical features, the absolute minimal pair and probing experiments show that BERT's knowledge is in fact not equal across these domains. We conclude that each method depicts different relevant aspects of a model's grammatical knowledge; comparing both gradient and absolute measures of performance of models gives a more complete picture. We recommend that future studies use multiple complementary methods to evaluate model performance.


\section{Related Work}

\paragraph{Evaluating Sentence Encoders} 
The success of sentence encoders and broader neural network methods in NLP has prompted significant interest in understanding the linguistic knowledge encapsulated in these models. 

A portion of related work focuses on Boolean classification tasks of English sentences to evaluate the grammatical knowledge encoded in these models. The objective of this task is to predict whether a single input sentence is acceptable or not, abstracting away from gradience in acceptability judgments. \newcite{linzen2016assessing} train classifiers on this task using data with manipulated verbal inflection to investigate whether LSTMs can identify subject-verb agreement violations, and therefore a (potentially long distance) syntactic dependency. \newcite{warstadt-2018} train models on this task using the CoLA corpus of acceptability judgments as a method for evaluating domain general grammatical knowledge, and \newcite{warstadt2019grammatical} analyze how these domain general classifiers perform on specific linguistic phenomena. \newcite{kann2019verb} use this task to evaluate whether sentence encoders represent information about verbal argument structure classes. 

A related method employs minimal pairs, consisting of two sentences that differ minimally in their content and differ in linguistic acceptability, to judge whether a model is sensitive to a single grammatical feature. \newcite{marvin2018targeted} and \citet{wilcox-2019} apply this method to phenomena such as subject-verb agreement, NPI licensing, and reflexive licensing.

Another branch of work uses probing tasks in which the objective is to predict the value of a particular linguistic feature given an input sentence. Probing tasks have been used to investigate whether sentence embeddings encode syntactic and surface features such as tense and voice \cite{shi2016does}, sentence length and word content \cite{adi2016fine}, or syntactic depth and morphological number \cite{conneau2018you}. \citet{giulianelli2018under} use diagnostic classifiers to track the propagation of information in RNN-based language models. \newcite{ettinger2018assessing} and \newcite{dasgupta2018evaluating} use automatic data generation to evaluate compositional reasoning. \citet{tenney2018what} introduce sub-sentence level probing tasks derived from common NLP tasks.



\paragraph{Negative Polarity Items} 
In the theoretical literature on NPIs, 
proposals have been made to unify the properties of the diverse NPI licensing environments. 
For example, a popular view states that NPIs are licensed if and only if they occur in downward entailing (DE)
environments \citep{fauconnier1975,ladusaw1979}, i.e. an environment that licences inferences from sets to subsets.\footnote{Other prominent theories of NPI licensing are based on notions of non-veridicality \citep{giannakidou1994,giannakidou1998,zwarts1998}, domain widening and emphasis \citep{kadmon-landman1993,krifka1995,chierchia2013}, a.o.} 

For instance, \ref{eg:de} shows that the environment under the scope of negation is DE. The property \emph{been to Paris} is a subset of \emph{been to France}, because those who have been to Paris are a subset of those who have been to France. While the inference from set to subset is normally invalid \ref{eg:de2}, it is valid if the property is embedded under negation \ref{eg:de1}. According to the DE theory of NPI licensing, this explains the contrast between \ref{gram} and \ref{ungram}.

\ex.\label{eg:de}
\a.\label{eg:de1} I haven't been to France.\\
$\rightarrow$ I haven't been to Paris. (DE)
\b.\label{eg:de2} I have been to France.
\\ $\not\rightarrow$ I have been to Paris. (not DE)

This view does not capture licensing in a number of environments, including  questions, the scope of \textit{only}, etc. No theory is yet accepted as identifying the unifying properties of all NPI licensing environments.

Within computational linguistics, NPIs are used as a testing ground for neural network models' grammatical knowledge. \newcite{marvin2018targeted} find that LSTM LMs do not systematically prefer sentences with licensed NPIs \ref{gram} over sentencew with unlicensed NPIs \ref{ungram}. \citet{jumelet2018language} shows LSTM LMs find a relation between the licensing
context and the negative polarity item, and appears to be aware of the scope of this context. \citet{wilcox-2019} use NPIs and filler-gap dependencies, as instances of non-local grammatical dependencies, to probe the effect of supervision with hierarchical structure. They find that structurally-supervised models outperform state-of-the-art sequential LSTM models, showing the importance of structure in learning non-local dependencies like NPI licensing.

\paragraph{CoLA}
We use the Corpus of Linguistic Acceptability \citep[CoLA;][]{warstadt-2018} in our experiments to train supervised acceptability classifiers. CoLA is a dataset of over 10k syntactically diverse example sentences from the linguistics literature with Boolean acceptability labels. As is conventional in theoretical linguistics, sentences are taken to be \textit{acceptable} if native speakers judge them to be possible sentences in their language. 
Such sentences are widely used in linguistics publications to illustrate phenomena of interest. The examples in CoLA are gathered from diverse sources and represent a wide array of syntactic, semantic, and morphological phenomena. As measured by the GLUE benchmark \citep{wang2018glue}, acceptability classifiers trained on top of BERT and related models reach near-human performance on CoLA.






\section{Methods}
\label{sec:methods}

We experiment with five approaches to the evaluation of grammatical knowledge of sentence representation models like BERT using our generated NPI acceptability judgment dataset (\cref{sec:data}). Each data sample in the dataset contains a sentence, a Boolean label which indicates whether the sentence is grammatically acceptable or not, and three Boolean meta-data variables (licensor, NPI, and scope; Table ~\ref{tb:2}). We evaluate four model types: BERT-large, BERT with fine-tuning on one of two tasks, and a simple bag-of-words baseline using GloVe word embeddings \citep{pennington2014glove}.



\paragraph{Boolean Acceptability}
We test the model's ability to judge the grammatical acceptability of the sentences in the NPI dataset. Following standards in linguistics, sentences for this task are assumed to be either totally acceptable or totally unacceptable. We fine-tune the sentence representation models to perform these Boolean judgments. For BERT-based sentence representation models, we add a classifier on top of the \texttt{[CLS]} embedding of the last layer. For BoW, we use a max pooling layer followed by an MLP classifier. The performance of the models is measured as Matthews Correlation Coefficient \cite[MCC;][]{matthews1975correlation}\footnote{MCC gives the correlation of two Boolean distributions between -1 and 1. A score of 0 is given to any two unrelated distributions, without requiring balanced class sizes.} between the predicted label and the gold label.

\paragraph{Absolute Minimal Pair}
We conduct a minimal pair experiment to analyze Boolean acceptability classifiers on minimally different sentences. 
Two sentences form a minimal pair if they differ in only one NPI-related Boolean meta-data variable within a paradigm, but have different acceptability. For example, the sentences in \ref{gram} and \ref{ungram} differ in whether an NPI licensor (negation) is present. We evaluate the models trained on acceptability judgments with the minimal pairs. In \textit{absolute minimal pair} evaluation, the models needs to correctly classify both sentences in the pair to be counted as correct.

\paragraph{Gradient Minimal Pair}
The \textit{gradient minimal pair} evaluation is a more lenient version of \textit{absolute minimal pair} evaluation: Here, we count a pair as correctly classified as long as the Boolean classifier's output for the acceptable class is higher for the acceptable sentence than for the unacceptable sentence. In other words, the classifier need only predict the acceptable sentence has the higher likelihood of being acceptable, but need not correctly predict that it is acceptable.

\paragraph{Cloze Test}
In the cloze test, a standard sentence-completion task, we use the masked language modeling (MLM) component in BERT \citep{devlin2018bert} and evaluate whether it assigns a higher probability to the acceptable sentence in a minimal pair, following \citet{linzen2016assessing}. An MLM predicts the probability of a single masked token based on the rest of the sentence. The minimal pairs tested are a subset of those in the absolute and gradient minimal pair experiments, where both sentences must be equal in length and differ in only one token. This differing token is replaced with \texttt{[MASK]}, and the minimal pair is taken to be classified correctly if the MLM assigns a higher probability to the token from the acceptable sentence. In contrast with the other minimal pair experiments, this experiment is entirely unsupervised, using BERT's native MLM functionality.

\paragraph{Feature Probing}
We use probing classifiers as a more fine-grained approach to the identification of grammatical variables. We freeze the sentence encoders both with and without fine-tuning from the acceptability judgment experiments and train lightweight classifiers on top of them to predict meta-data labels corresponding to the key properties a model must learn in order to judge the acceptability of NPI sentences. These properties are whether a licensor is present, whether an NPI is present, and whether the NPI (or a similar item) is in the syntactic scope of the licensor (or a similar item). Crucially, each individual meta-data label by itself does not decide acceptability (i.e., these probing experiments test a different but related set of knowledge from acceptability experiments).

\begin{table*}[!th]
\begin{center}
\footnotesize
\begin{tabular}{lll}
\toprule
 Environment & Abbrev. & Example \\
\midrule
 Adverbs & ADV & The guests who \textbf{rarely} love \textit{any} actors had left libraries. \\ 
 Conditionals & COND & \textbf{If} the pedestrian passes \textit{any} schools, the senator will talk to the adults. \\
 Determiner negation & D-NEG & Just as the waitress said, \textbf{no} customers thought that \textit{any} dancers bought the dish. \\
 Sentential negation & S-NEG & These drivers have \textbf{not} thought that \textit{any} customers have lied. \\
 Only & ONLY & From what the cashier heard, \textbf{only} the children have known \textit{any} dancers. \\
 Quantifiers & QNT & \textbf{Every} actress who was talking about \textit{any} high schools criticizes the children. \\
 Questions & QUES & The boys wonder \textbf{whether} the doctors went to \textit{any} art galleries. \\
 Simple questions & SMP-Q & \textbf{Has} the guy worked with \textit{any} teenagers\textbf{?} \\
 Superlatives & SUP & The teenagers approach the \textbf{nicest} actress that \textit{any} customers had criticized. \\
\bottomrule
\end{tabular}
\end{center}
\caption{Examples of each of the NPI licensing environments generated. The licensor is in bold, and the NPI (here \textit{any}) is in italics. All examples show cases where the NPI is present (NPI=1), the licensor is present (Licensor=1), and the NPI is in the scope of the licensor (Scope=1); all are acceptable to native speakers of English.}
\label{tb:1}
\end{table*}

\section{Data}\label{sec:data}

In order to probe BERT's performance on sentences involving NPIs, we generate a set of sentences and acceptability labels for the experiments in this paper. We use generated data so that we can assess minimal pairs, and so that there are sufficient unacceptable sentences. We release all our data\footnote{\href{https://alexwarstadt.files.wordpress.com/2019/08/npi_lincensing_data.zip}{\url{https://alexwarstadt.files.wordpress.com/2019/08/npi_lincensing_data.zip}} (Clicking on the link will cause a file to download.)} and our generation code and vocabulary.\footnote{\href{https://github.com/alexwarstadt/data_generation}{\url{https://github.com/alexwarstadt/data\_generation}}}

\begin{table*}[!h]
\begin{center}
\footnotesize
\begin{tabular}{cccl}
\toprule
 Licensor & NPI & Scope & Sentence \\
\midrule
\tikzmark{111} 1 & 1 & 1 & Those boys wonder \textbf{whether} [the doctors \textit{ever} went to an art gallery.] \\ 
\tikzmark{110} 1 & 1 & 0 & {*}Those boys \textit{ever} wonder \textbf{whether} [the doctors went to an art gallery.] \\
\tikzmark{101} 1 & 0 & 1 & Those boys wonder \textbf{whether} [the doctors \textit{often} went to an art gallery.] \\ 
\tikzmark{100} 1 & 0 & 0 & Those boys \textit{often} wonder \textbf{whether} [the doctors went to an art gallery.] \\
\tikzmark{011} 0 & 1 & 1 & {*}Those boys say \textbf{that} [the doctors \textit{ever} went to an art gallery.] \\
\tikzmark{010} 0 & 1 & 0 & {*}Those boys \textit{ever} say \textbf{that} [the doctors went to an art gallery.] \\
\tikzmark{001} 0 & 0 & 1 & Those boys say \textbf{that} [the doctors \textit{often} went to an art gallery.] \\
\tikzmark{000} 0 & 0 & 0 & Those boys \textit{often} say \textbf{that} [the doctors went to an art gallery.] \\
\bottomrule
\end{tabular}
\begin{tikzpicture}[remember picture, overlay]
  \draw ($(pic cs:010) + (-1.5em,0.25em)$) edge[bend right=60,-stealth] ($(pic cs:000) + (-1.5em,0.25em)$);
  \draw ($(pic cs:011) + (-1.5em,0.25em)$) edge[bend right=60,-stealth] ($(pic cs:001) + (-1.5em,0.25em)$);
  \draw ($(pic cs:110) + (-1.5em,0.25em)$) edge[bend right=60,-stealth] ($(pic cs:100) + (-1.5em,0.25em)$);
  \draw ($(pic cs:011) + (-1.5em,0.25em)$) edge[bend left=60,-stealth] ($(pic cs:111) + (-1.5em,0.25em)$);
  \draw ($(pic cs:110) + (-1.5em,0.25em)$) edge[bend left=60,-stealth] ($(pic cs:111) + (-1.5em,0.25em)$);
\end{tikzpicture}
\end{center}
\caption{Example 2$\times$2$\times$2 paradigm using the Questions environment. The licensor (\textit{whether}) or licensor replacement (\textit{that}) is in bold. The NPI (\textit{ever}) or NPI replacement (\textit{often}) is in italics. When licensor=1, the licensor is present rather than its replacement word. When NPI=1, the NPI is present rather than its replacement. The scope of the licensor/licensor replacement is shown in square brackets (brackets, italicization, and boldface are not present in the actual data). When scope=1, the NPI/NPI replacement is within the scope of the licensor/licensor replacement. Unacceptable sentences are marked with {*}. The five minimal pairs are connected by arrows that point from the unacceptable to the acceptable sentence.}
\label{tb:2}
\end{table*}

\paragraph{Licensing Features}

We create a controlled set of 136,000 English sentences using an automated sentence generation procedure, inspired in large part by previous work by \citet{ettinger2016probing,ettinger2018assessing}, \citet{marvin2018targeted}, \citet{dasgupta2018evaluating}, and \citet{kann2019verb}. The set contains nine NPI licensing environments (Table \ref{tb:1}), and two NPIs (\textit{any}, \textit{ever}). All but one licensor-NPI pair follows a 2$\times$2$\times$2 paradigm, which manipulates three boolean NPI licensing features: licensor presence, NPI presence, and the occurrence of an NPI within a licensor's scope. Each 2$\times$2$\times$2 paradigm forms 5 minimal pairs. Table \ref{tb:2} shows an example paradigm.

\begin{figure*}[!h]
  \centering
  \includegraphics[width=0.9\linewidth]{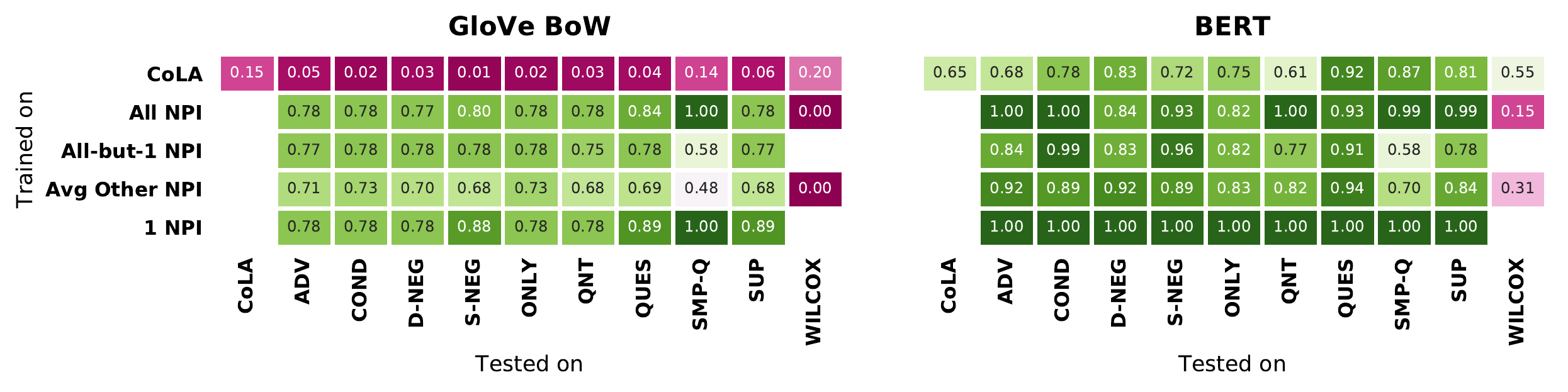}
  \caption{\label{fig:main_exp} Results from the acceptability judgment experiment in MCC. The columns indicate evaluation tests, and the rows fine-tuning settings.}
\end{figure*}

The \emph{Licensor} feature indicates whether an NPI licensor is present in the sentence. For many environments, there are multiple lexical items that serve as a licensor (e.g., the adverbs environment contains \textit{rarely}, \textit{hardly}, \textit{seldom}, \textit{barely}, and \textit{scarcely} as NPI licensors). When the licensor is not present, we substitute it with a \emph{licensor replacement} that has a similar syntactic distribution but does not license NPIs, again using multiple appropriate lexical items as replacements. For example, licensor replacements for quantifier licensors like \emph{every} include quantifiers like \textit{some}, \textit{many}, and \textit{more than three} that do not license NPIs.

The \emph{NPI} feature indicates whether an NPI is in the sentence or if it is substituted by an \emph{NPI replacement} with similar structural distribution. For example, NPI replacements for \emph{ever} include other adverbs such as \emph{often}, \emph{sometimes}, and \emph{certainly}.

The \emph{Scope} feature indicates whether the NPI/NPI replacement is within the scope of the licensor/licensor replacement. As illustrated earlier in \ref{nosc}, a sentence containing an NPI is only acceptable when the NPI falls within the scope of the licensor. What constitutes the scope of a licensor is highly dependent on the type of licensor.


The exception to the 2$\times$2$\times$2 paradigm is the Simple Questions licensing condition, with a reduced 2$\times$2 paradigm. It lacks a scope manipulation because the question takes scope over the entire clause, and in Simple Questions the clause is the whole sentence. The paradigm for Simple Questions is given in Table \ref{tb:3_appendix} in the Appendix; it forms only 2 minimal pairs.

\paragraph{Data Generation}
To generate the sentences, we create sentence templates for each paradigm. Templates follow the general structure illustrated in example \ref{template}, in which the part-of-speech (auxiliary verb, determiner, noun, verb), as well as the instance number is specified. For example, N2 is the second instance of a noun in the template. We use these labels here for illustrative purposes; in reality, the templates also include more fine-grained specifications, such as verb tense and noun number. 

\exg.\label{template} Aux1 D1 N1 V1 any N2 ? \\ {} Has the guy {seen} any waitresses ? \\

Given the specifications encoded in the sentence templates, words were sampled from a vocabulary of over 1000 lexical items annotated with 30 syntactic, morphological, and semantic features. The annotated features allow us to encode selectional requirements of lexical items, e.g., what types of nouns a verb can combine with. This avoids blatantly implausible sentences. 

For each environment, the training set contains 10K sentences, and the dev and test sets contain 1K sentences each. Sentences from the same paradigm are always in the same set. 

In addition to our data set, we also test BERT on a set of 104 handcrafted sentences from the NPI sub-experiment in \citet{wilcox-2019}, who use a paradigm that partially overlaps with ours, but has an additional condition where the NPI linearly follows its licensor while not being in the scope of the licensor. This is included as an additional test set for evaluating acceptability classifiers in (\ref{result:accept}).

\paragraph{Data validation}
We use Amazon Mechanical Turk (MTurk) to validate a subset of our sentences to assure that the generated sentences represent a real contrast in acceptability.
We randomly sample five-hundred sentences from the dataset, sampling approximately equally from each environment, NPI and paradigm.
Each sentence is rated on a Likert scale of 1-6, with 1 being ``the sentence is not possible in English'' and 6 being ``the sentence is possible in English'' by 20 unique participants located in the US who self-identified as native English speakers. Participants are compensated \$0.25 per HIT and are often able to complete a HIT of 5 sentences in just under 1 minute.

Table \ref{tab:mturk_validation} in the Appendix shows the participants' scores transformed into a Boolean judgment of 0 (unacceptable, score $\leq$ 3) or 1 (acceptable, score $\geq$ 4) and presented as the percentage of `acceptable' ratings assigned to the sentences in each of the NPI-licensing environments.
Across all NPI-licensing environments, 81.3\% of the sentences labelled acceptable are assigned an acceptable rating by the MTurk raters, and 85.2\% of sentences labeled unacceptable are assigned an unacceptable rating.
This gives an overall agreement rating of 82.8\%\footnote{We observe that sentences with `any' get over-accepted. Overall agreement for sentences with `ever' is 87.3\%, while agreement for those with `any' is 78.3\%. We believe this is due to a free-choice interpretation of `any' occurring more easily than is often reported in the semantics literature.}.
A Wilcoxon signed-rank test \citep{wilcoxon1945} shows that within each environment and for each NPI, the acceptable sentences are more often rated as acceptable by our MTurk validators than the unacceptable sentences (all \textit{p}-values \textless\ 0.001).
This contrast holds considering both the raw Likert-scale responses and the responses transformed to a Boolean judgment.

\section{Experimental Settings}

We conduct our experiments with the jiant 0.9 \footnote{\href{https://github.com/nyu-mll/jiant/tree/blimp-and-npi/scripts/bert_npi}{\url{https://github.com/nyu-mll/jiant/tree/blimp-and-npi/scripts/bert_npi}}} \citep{wang2019jiant} multitask learning and transfer learning toolkit, the AllenNLP platform \citep{gardner2018allennlp}, and the BERT implementation from HuggingFace.\footnote{\url{https://github.com/huggingface/pytorch-pretrained-BERT}}




\paragraph{Models}\label{sec:models}

We study the following sentence understanding models: (i) \newterm{GloVe BoW}: a bag-of-words baseline obtained by max-pooling of 840B tokens 300-dimensional GloVe word embeddings \cite{pennington2014glove} and (ii) \newterm{BERT} \cite{devlin2018bert}: we use the cased version of BERT-large model, which works the best for our tasks in pilot experiments. 
In addition, since recent work \citep{liu2019mtdnn,stickland2018pals} has shown that intermediate training on related tasks can meaningfully impact BERT's performance on downstream tasks, we also explore two additional BERT-based models---(iii) \newterm{BERT$\rightarrow$MNLI}: BERT fine-tuned on the Multi-Genre Natural Language Inference corpus \cite{williams2018broad}, motivated both by prior work on pretraining sentence encoders on MNLI \citep{conneau2017supervised} as well as work showing significant improvements to BERT on downstream semantic tasks \cite{phang2018stilts, bowman2018looking} (iv) \newterm{BERT$\rightarrow$CCG}: BERT fine-tuned on Combinatory Categorial Grammar Bank corpus \citep{hockenmaier2007ccgbank}, motivated by \citeauthor{wilcox-2019}'s (\citeyear{wilcox-2019}) finding that structural supervision may improve a LSTM-based sentence encoder’s knowledge on non-local syntactic dependencies.

\begin{figure*}[!h]
  \centering
  \includegraphics[width=0.9\linewidth]{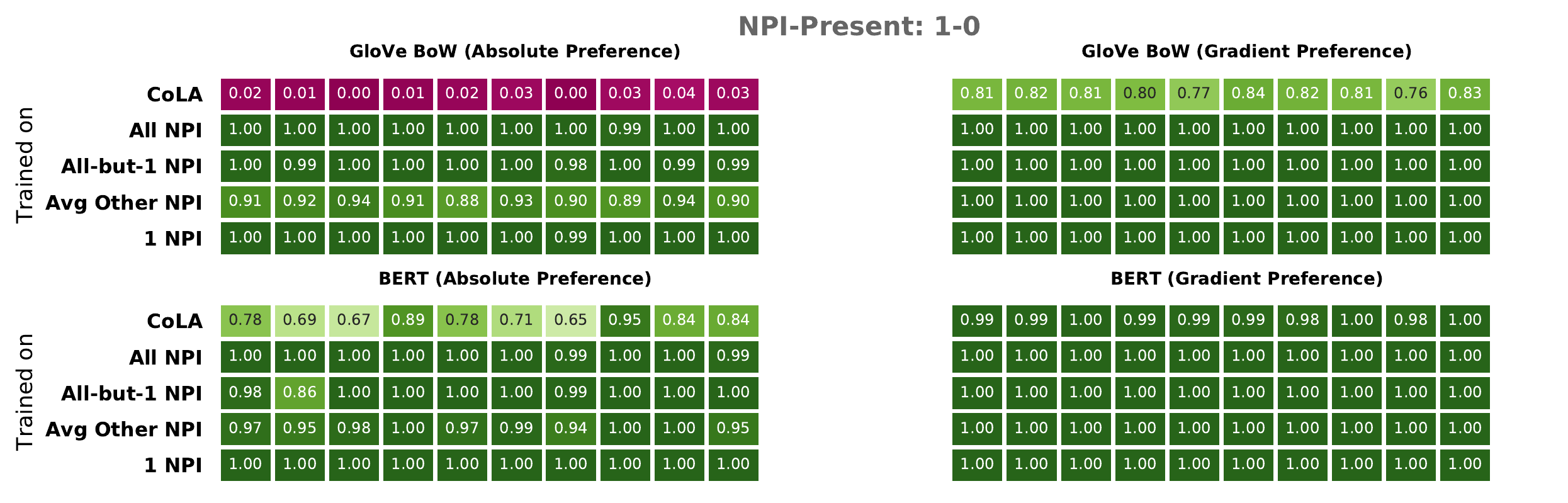}
  \includegraphics[width=0.9\linewidth]{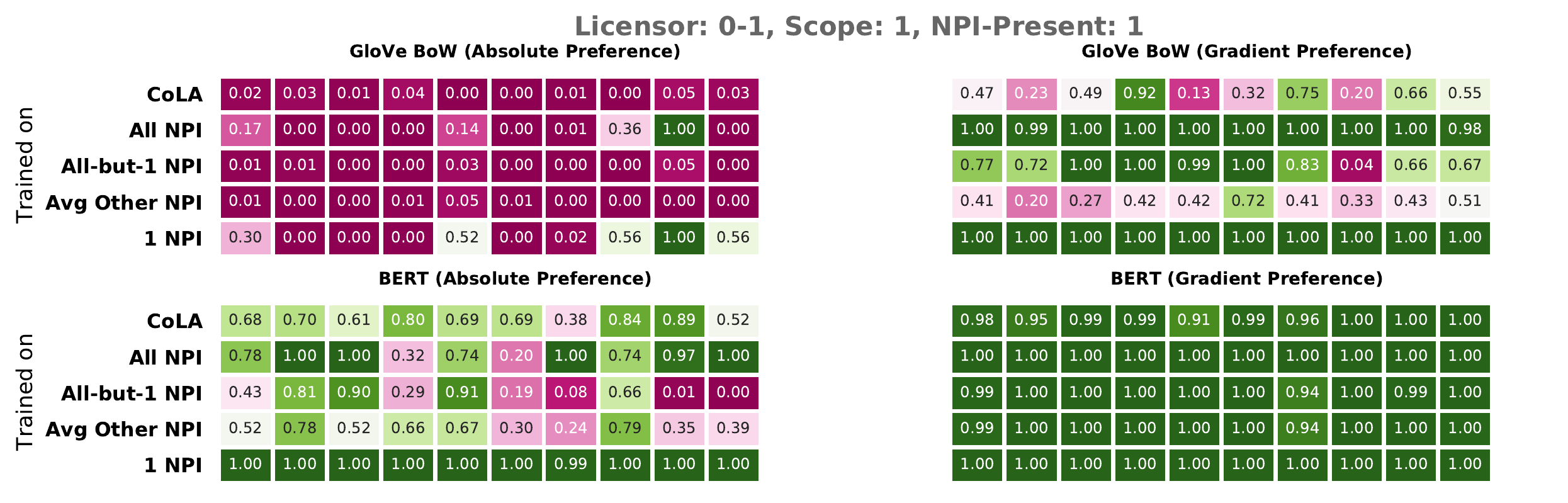}
  \includegraphics[width=0.9\linewidth]{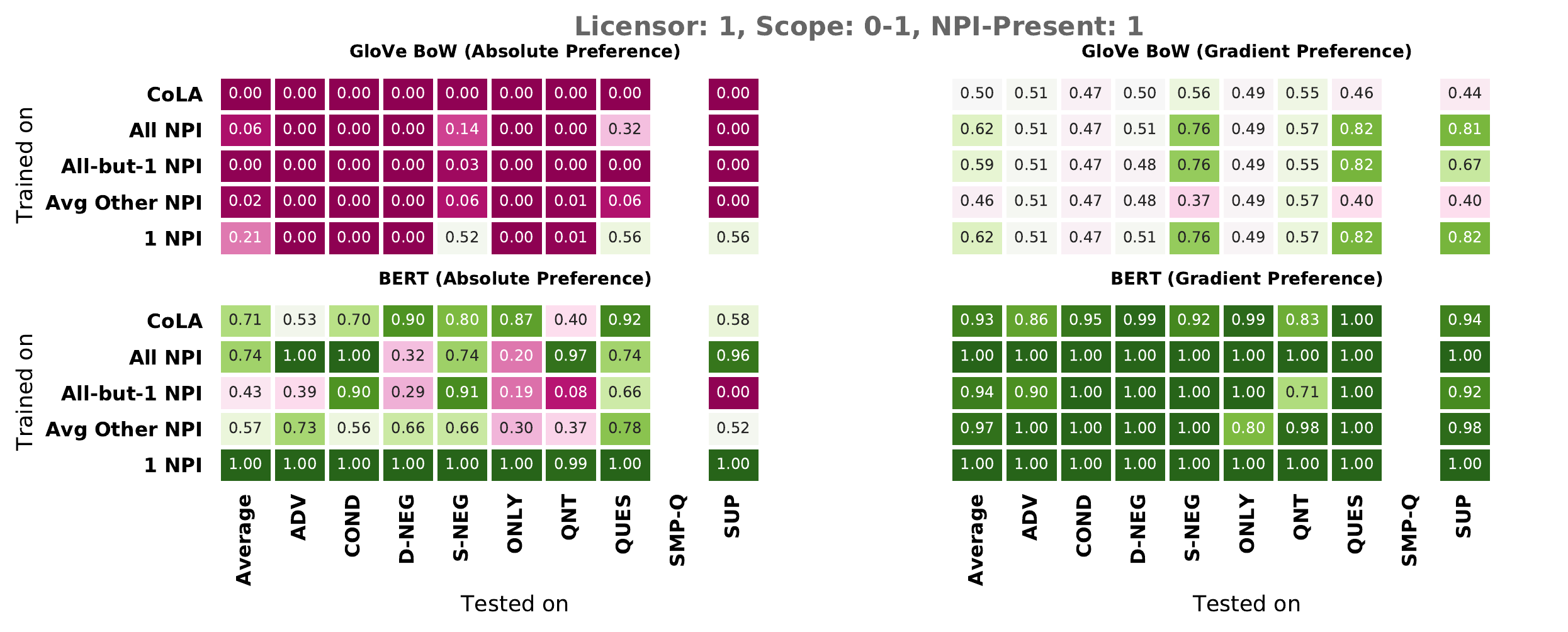}
  \caption{\label{fig:minpairs} Results from the minimal pair test. The top section shows the average accuracy for detecting the presence of the NPI, the middle section shows average accuracy for detecting the presence of the licensor, and the bottom shows average accuracy of minimal pair contrasts that differ in whether the NPI is in scope of the licensor. Within each section, we show performance of GloVe BoW and BERT models under both absolute preference and gradient preference evaluation methods. The rows represent the training-evaluation configuration, while the columns represent different licensing environments.}
\end{figure*}

\paragraph{Training-Evaluation Configurations}\label{sec:traintoeval}

We are interested in whether sentence representation models learn NPI licensing as a unified property. Can the models generalize from trained environments to previously unseen environments? To answer these questions, for each NPI environment, we extensively test the performance of the models in the following configurations: (i) \newterm{CoLA}: training on CoLA, evaluating on the environment. (ii) \newterm{1 NPI}: training and evaluating on the same NPI environment. (iii) \newterm{Avg Other NPI}: training independently on every NPI environment except one, averaged over the evaluation results on that environment. (iv) \newterm{All-but-1 NPI}: training on all environments except for one environment, evaluating on that environment. (v) \newterm{All NPI}: training on all environments, evaluating on the environment.

\section{Results}\label{sec:results}






\paragraph{Acceptability Judgments}\label{result:accept}

The results in Fig.\,\ref{fig:main_exp} show that BERT  outperforms the BoW baseline on all test data with all fine-tuning settings. Within each BERT variants, MCC reaches 1.0 on all test data in the \newterm{1 NPI} setting. When the \newterm{All-but-1 NPI} training-evaluation configuration is used, the performance on all NPI environments for BERT drops. While the MCC value on environments like conditionals and sentential negation remains above 0.9, on the simple question environment it drops to 0.58. Compared with NPI data fine-tuning, CoLA fine-tuning results in BERT's lower performance on most of the NPI environments but better performance on data from \citet{wilcox-2019}. 


In comparing the three BERT variants (see full results in Figure \ref{fig:main_exp_appendix} in the Appendix), the \newterm{Avg Other NPI} shows that on 7 out of 9 NPI environments, plain BERT outperforms BERT$\to$MNLI and BERT$\to$CCG. Even in the remaining two environments, plain BERT yields about as good performance as BERT$\to$MNLI and BERT$\to$CCG, indicating that MNLI and CCG fine-tuning brings no obvious gain to acceptability judgments.


\paragraph{Absolute and Gradient Minimal Pairs}\label{minimal-pair}

The results (Fig.\,\ref{fig:minpairs}) show that models' performance hinges on how minimal pairs differ. When tested on minimal pairs differing by the presence of an NPI, BoW and plain BERT obtain (nearly) perfect accuracy on both absolute and gradient measures across all settings. For minimal pairs differing by licensor and scope, BERT again achieves near perfect performance on the gradient measure, while BoW does not. On the absolute measure, both BERT and BoW perform worse. Overall, it shows that absolute judgment is more challenging when targeting licensor, which involves a larger pool of lexical items and syntactic configurations than NPIs; and scope, which requires nontrivial syntactic knowledge about NPI licensing.

As in the acceptability experiment, we find that intermediate fine-tuning on MNLI and CCG does not improve performance (see full results in Figures \ref{fig:minpairs_appendix1}-\ref{fig:minpairs_appendix3} in Appendix).

\begin{figure}[!h]
  \centering
  \includegraphics[width=0.85\linewidth]{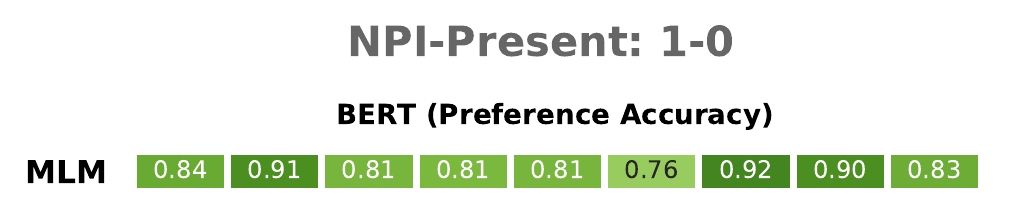}
  \includegraphics[width=0.85\linewidth]{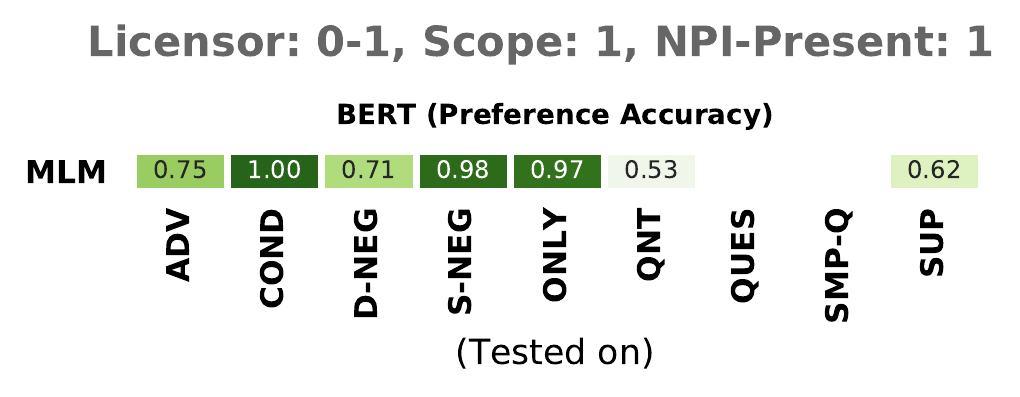}
  \caption{\label{fig:mlm} Results of BERT MLM performance in the cloze test. The top section shows the average accuracy for detecting the presence of the NPI; the bottom section show the accuracy for detecting the presence of the licensor. The columns represent different licensing environments}
\end{figure}

\paragraph{Cloze Test}

The results (Fig.\,\ref{fig:mlm}) show that even without supervision on NPI data, the BERT MLM can distinguish between acceptable and unacceptable sentences in the NPI domain. Performance is highly dependent on the NPI-licensing environment and type of minimal pair. Accuracy for detecting NPI presence falls between $0.76$ and $0.93$ for all environments. Accuracy for detecting licensor presence is much more variable, with the BERT MLM achieving especially high performance in conditional, sentential negation, and \emph{only} environments; and low performance in quantifier and superlative environments.

\paragraph{Feature Probing} \label{sec:probing}

\begin{figure*}[!h]
  \centering
  \includegraphics[width=\textwidth]{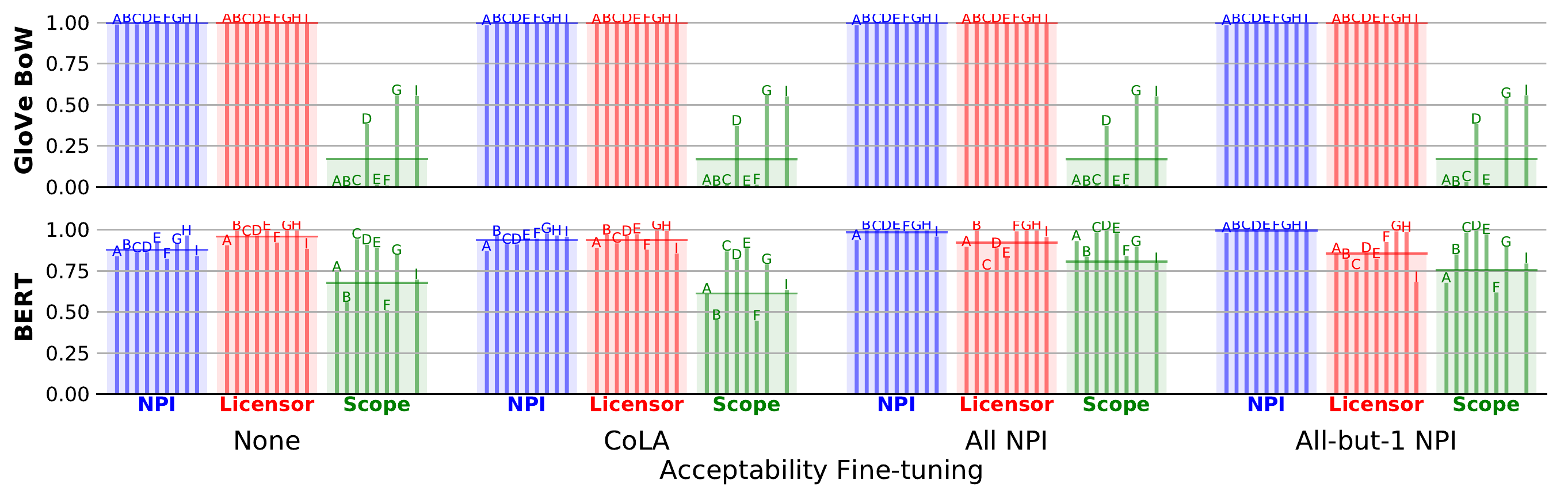}
  \caption{Results of probing classification on NPI presence, licensor presence, and whether the NPI is in the scope of the licensor (scope detection), shown in MCC. Letters on top of bars refer to NPI environments: A=ADV, B=COND, C=D-NEG, D=S-NEG, E=ONLY, F=QNT, G=QUES, H=SMP-Q, I=SUP.}
  \label{probing}
\end{figure*}



Results (Fig.\,\ref{probing}) show that plain BERT outperforms the BoW baseline in detecting whether NPI is in the scope of the licensor (henceforth `scope detection'). As expected, BoW is nearly perfect in detecting presence of NPI and licensor, as these tasks do not require knowledge of syntax or word order. Consistent with results from previous experiments, detecting the presence of a licensor is slightly more challenging for models fine-tuned with CoLA or NPI data. However, the overall lower performances in scope detection compared with detecting the presence of the licensor is not found in the minimal-pair experiments. 


CoLA fine-tuning improves the performance for BERT, especially for detecting NPI presence. Fine-tuning on NPI data improves scope detection. Inspection of environment-specific results shows that models struggle when the superlative, quantifiers, and adverb environments are the held-out test sets in the All-but-1 NPI fine-tuning setting.

Different from other experiments, BERT and BERT$\to$MNLI have comparable performance across many settings and tasks, beating BERT$\to$CCG especially in scope detection (see full results in Figure \ref{probing_appendix} in the Appendix).

\section{Discussion}

We find that BERT systematically represents all features relevant to NPI licensing across most individual licensing environments, according to certain evaluation methods. However, these results vary widely across the different methods we compare. In particular, BERT performs nearly perfectly on the gradient minimal pairs task across all of minimal pair configurations and nearly all licensing environments. Based on this method alone, we might conclude that BERT's knowledge of this domain is near perfect. However, the other methods show a more nuanced picture.

BERT's knowledge of which expressions are NPIs and NPI licensors is generally stronger than its knowledge of the licensors' scope. This is especially apparent from the probing results (Fig.\,\ref{probing}). BERT without acceptability fine-tuning performs close to ceiling on detecting the presence of a licensor, but is inconsistent at scope detection. Tellingly, the BoW baseline is also able to perform at ceiling on telling whether a licensor is present. For BoW to succeed at this task, the GloVe embeddings for NPI-licensors must share some common property, most likely the fact that licensors co-occur with NPIs. It is possible that BERT is able to succeed using a similar strategy. By contrast, identifying whether an NPI is in the scope of a licensor requires at the very least word order information and not just co-occurrences.


The contrast in BERT's performance on the gradient and absolute tasks tells us that these evaluations reveal different aspects of BERT's knowledge. The gradient task is strictly easier than the absolute task. On the one hand, BERT's high performance on the gradient task reveals the presence of systematic knowledge in the NPI domain. On the other hand, due to ceiling effects, the gradient task fails to reveal actual differences between NPI-licensing environments that we clearly observe based on absolute, cloze, and probing tasks.

While BERT has systematic knowledge of acceptability contrasts, this knowledge varies across different licensing environments and is not categorical. Generative models of syntax \citep{chomsky1965aspects,chomsky1981lectures,chomsky1995minimalist} model human knowledge of natural language as categorical: In that sense BERT fails at attaining human performance. However, some have argued that acceptability is inherently gradient \citep{lau2017grammaticality}, and results from the human validation study on our generated dataset show evidence of gradience in the acceptability of sentences in our dataset.




Supplementing BERT with additional pretraining on CCG and MNLI does not improve performance, and even lowers performance in some cases. While results from \citet{phang2018stilts} lead us to hypothesize that intermediate pretraining might help, this is not what we observe on our data. This result is in direct contrast with the results from \citet{wilcox-2019}, who find that syntactic pretraining does improve performance in the NPI domain. This difference in findings is likely due to differences in models and training procedure, as their model is an RNN jointly trained on language modeling and parsing over the much smaller Penn Treebank \cite{marcus1993building}.

Future studies would benefit from employing a variety of different methodologies for assessing model performance withing a specified domain. In particular, a result showing generally good performance for a model should be regarded as possibly hiding actual differences in performance that a different task would reveal. Similarly, generally poor performance for a model does not necessarily mean that the model does not have systematic knowledge in a given domain; it may be that an easier task would reveal systematicity.

\section{Conclusion}
We have shown that within a well-defined domain of English grammar, evaluation of sentence encoders using different tasks will reveal different aspects of the encoder's knowledge in that domain. By considering results from several evaluation methods, we demonstrate that BERT has systematic knowledge of NPI licensing. However, this knowledge is unequal across the different features relevant to this phenomenon, and does not reflect the Boolean effect that these features have on acceptability.



 
\section*{Acknowledgments}

This project was a joint effort by the participants in the Spring 2019 NYU Linguistics seminar course \textit{Linguistic Knowledge in Reusable Sentence Encoders}. We are grateful to the department for making this seminar possible.

This material is based upon work supported by the National Science Foundation under Grant No. 1850208. Any opinions, findings, and conclusions or recommendations expressed in this material are those of the author(s) and do not necessarily reflect the views of the National Science Foundation. This project has also benefited from financial support to SB by Samsung Research under the project \textit{Improving Deep Learning using Latent Structure}
and from the donation of a Titan V GPU by NVIDIA Corporation. 


\bibliography{emnlp-ijcnlp-2019}
\bibliographystyle{acl_natbib}

\clearpage
\section*{Appendix}
\begin{figure*}[!h]
  \centering
  \includegraphics[width=0.85\linewidth]{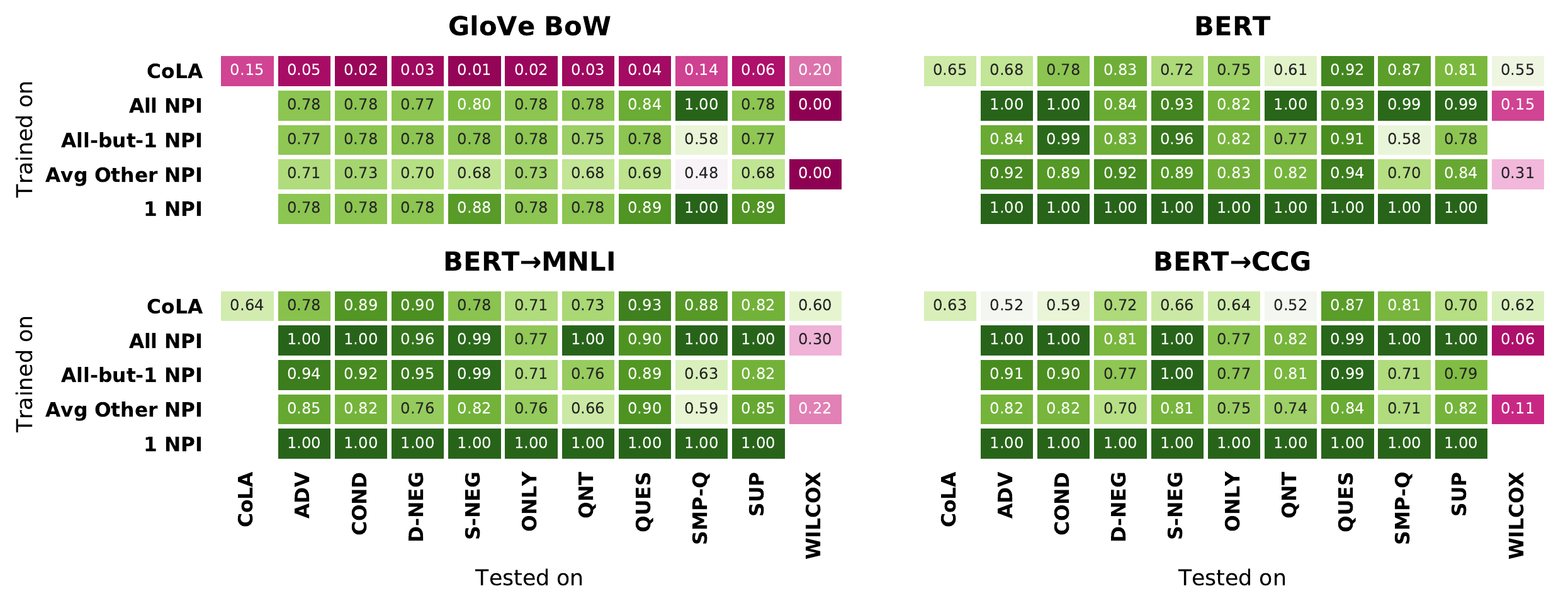}
  \caption{\label{fig:main_exp_appendix} Results from the acceptability judgment experiment in MCC. The columns indicate evaluation tests, and the rows fine-tuning settings.}
\end{figure*}

\begin{figure*}[!h]
  \centering
  \includegraphics[width=0.85\linewidth]{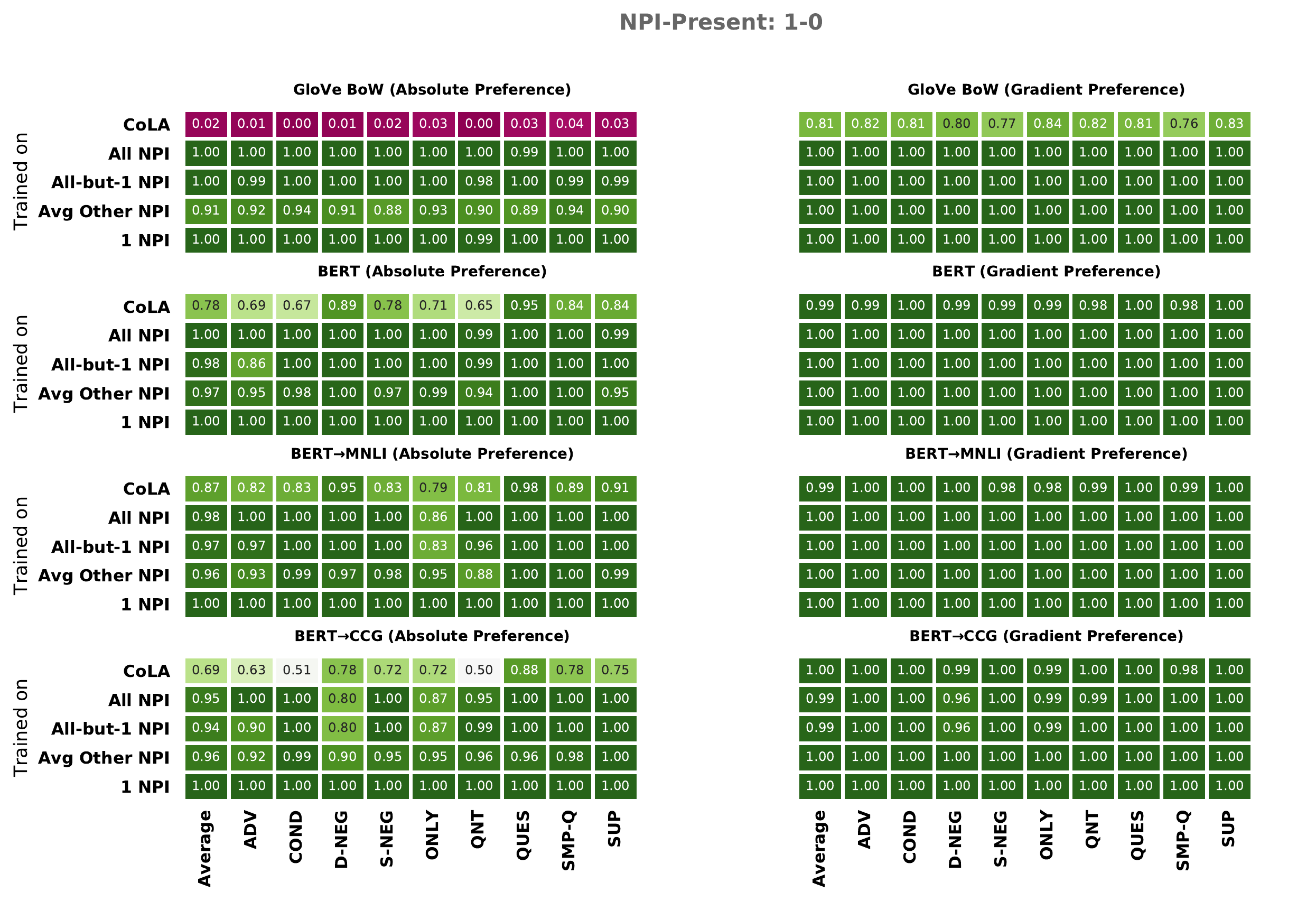}
  \caption{\label{fig:minpairs_appendix1} Results from minimal pair test for the NPI-presence contrast. The smaller diagrams of each sector show performance of BoW and BERT variants under two different minimal pair evaluation methods. The rows represent training-evaluation configuration, while the columns represent different licensing environments.}
\end{figure*}
\begin{figure*}[!h]
  \centering
  \includegraphics[width=0.85\linewidth]{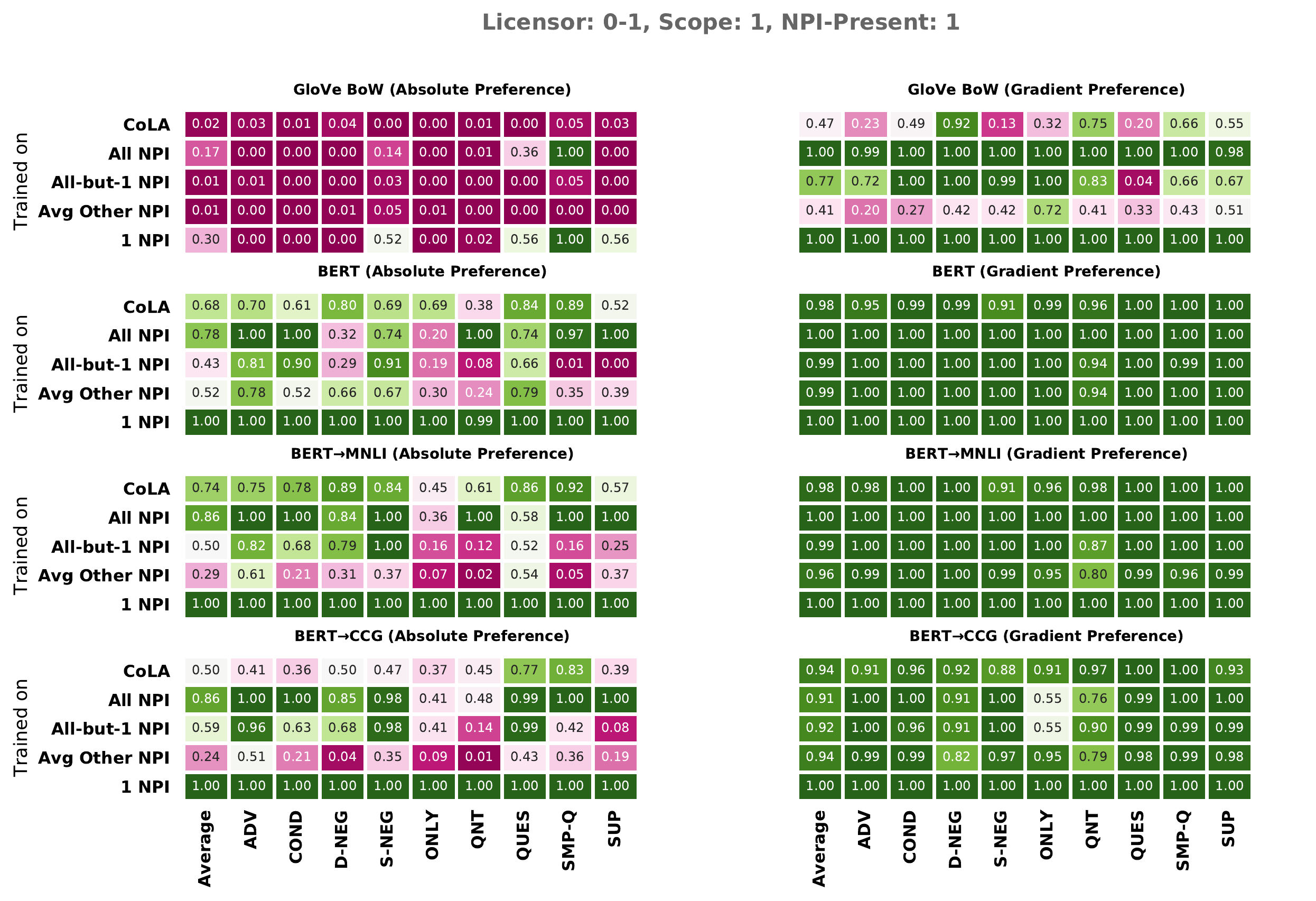}
  \caption{\label{fig:minpairs_appendix2} Results from minimal pair test for the licensor-presence contrast. The smaller diagrams of each sector show performance of BoW and BERT variants under two different minimal pair evaluation methods. The rows represent training-evaluation configuration, while the columns represent different licensing environments.}
\end{figure*}
\begin{figure*}[!h]
  \centering
  \includegraphics[width=0.85\linewidth]{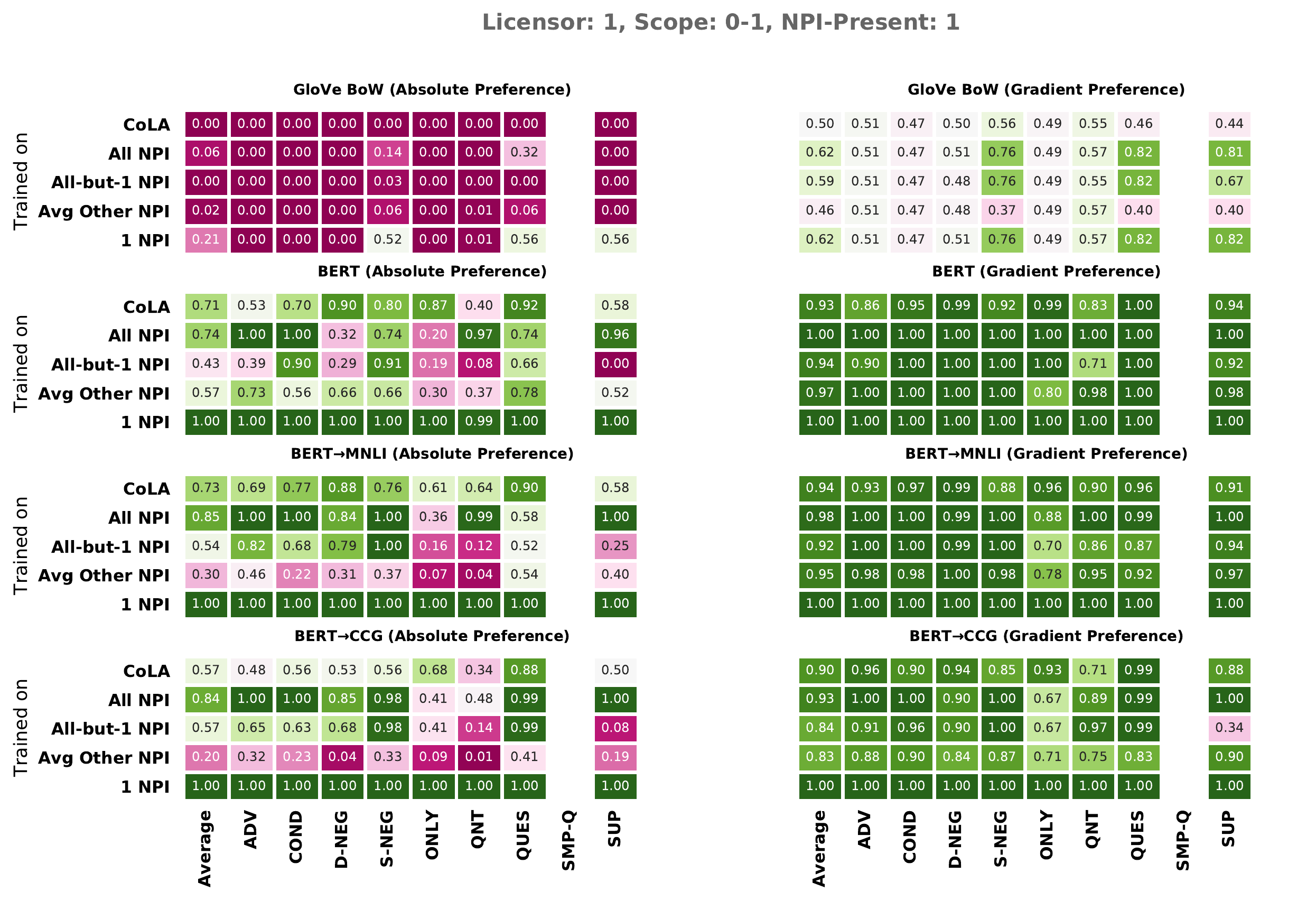}
  \caption{\label{fig:minpairs_appendix3} Results from minimal pair test for the scope contrast. The smaller diagrams of each sector show performance of BoW and BERT variants under two different minimal pair evaluation methods. The rows represent training-evaluation configuration, while the columns represent different licensing environments.}
\end{figure*}

\begin{figure*}[!h]
  \centering
  \includegraphics[width=\textwidth]{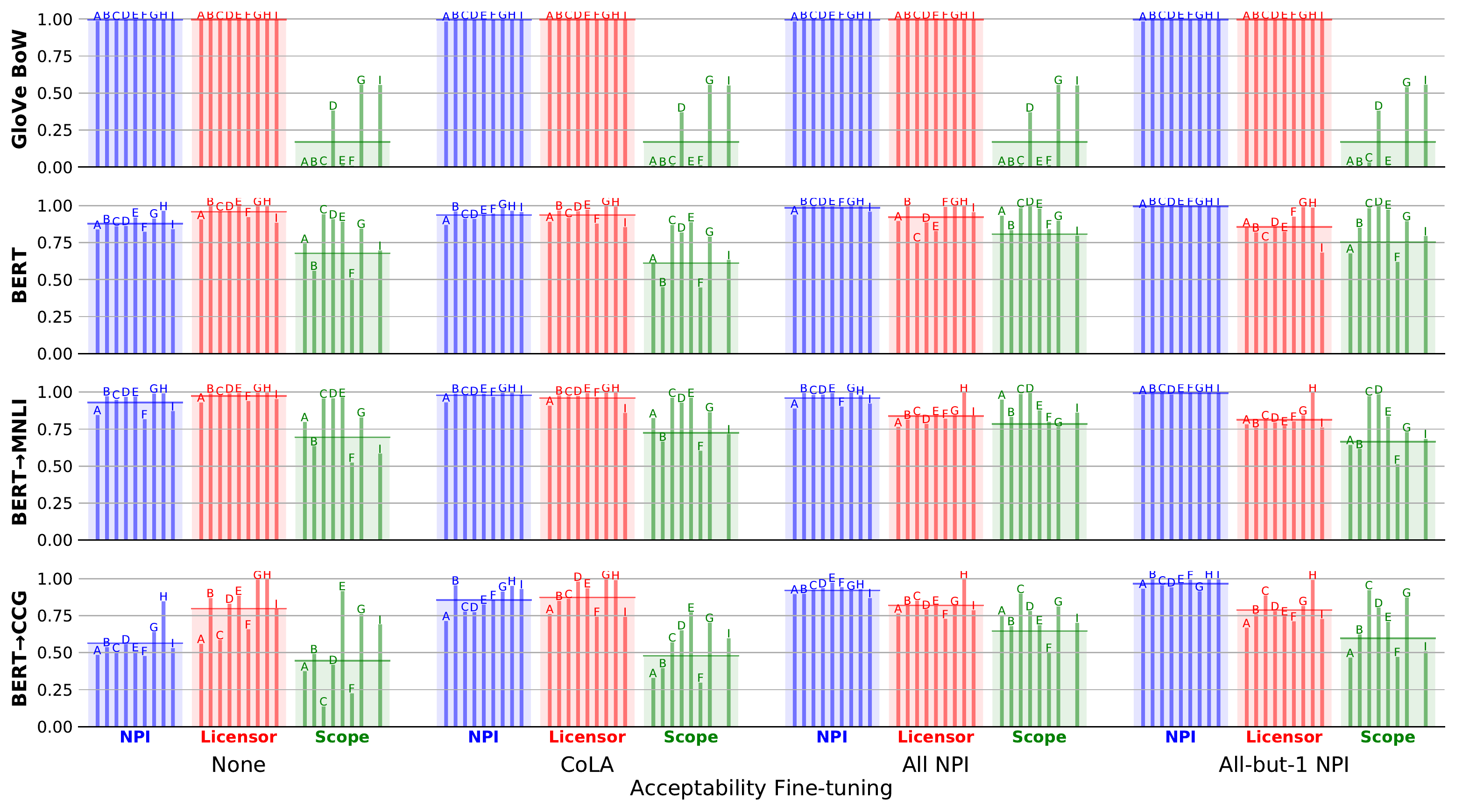}
  \caption{Results of probing classification on NPI presence, licensor presence, and scope detection, shown in MCC. Letters on top of bars refer to NPI environments: A=ADV, B=COND, C=D-NEG, D=S-NEG, E=ONLY, F=QNT, G=QUES, H=SMP-Q, I=SUP.}
  \label{probing_appendix}
\end{figure*}

\begin{table}[!h]
\begin{center}
\footnotesize
\begin{tabular}{ccl}
\toprule
 Lic. & NPI & Sentence \\
\midrule
\tikzmark{11} 1 & 1 & \textbf{Has} the guy worked with \textit{any} teenagers? \\ 
\tikzmark{10} 1 & 0 & \textbf{Has} the guy worked with \textit{the} teenagers? \\
\tikzmark{01} 0 & 1 & {*}The guy \textbf{has} worked with \textit{any} teenagers. \\ 
\tikzmark{00} 0 & 0 & The guy \textbf{has} worked with \textit{the} teenagers. \\
\bottomrule
\end{tabular}
\begin{tikzpicture}[remember picture, overlay]
  \draw ($(pic cs:01) + (-1.5em,0.25em)$) edge[bend right=60,-stealth] ($(pic cs:00) + (-1.5em,0.25em)$);
  \draw ($(pic cs:01) + (-1.5em,0.25em)$) edge[bend left=60,-stealth] ($(pic cs:11) + (-1.5em,0.25em)$);
\end{tikzpicture}
\end{center}
\caption{Reduced paradigm for Simple questions. ``Lic.'' is abbreviated from ``Licensor''. The licensor and licensor replacement are shown in bold (\textit{has} in both cases). The NPI (\textit{any}) and NPI replacement (\textit{the}) are shown in italics. There is no scope manipulation because it is not possible to place an NPI or NPI replacement outside of the scope of an interrogative or declarative phrase. The 2 minimal pairs are shown by arrows, pointing from unacceptable to acceptable  sentence.}
\label{tb:3_appendix}
\end{table}

\begin{table}[h!]
    \centering \small
    \begin{tabular}{p{0.25\linewidth} c r r}
        \toprule
        Environment & Label & \% accept & Diff \\
        \midrule
        \multirow{2}{0.25\linewidth}{Adverb} & * & 8.33 & \multirow{2}{*}{61.67} \\ 
        & {\checkmark} & 70.00 & \\
        \midrule
        \multirow{2}{0.25\linewidth}{Conditionals} & * & 37.50 & \multirow{2}{*}{50.00}\\ 
        & {\checkmark} & 87.50 & \\
        \midrule
        \multirow{2}{0.25\linewidth}{Determiner negation} & * & 11.11 & \multirow{2}{*}{78.89}\\ 
        & {\checkmark} & 90.00 &\\
        \midrule
        \multirow{2}{0.25\linewidth}{Embedded questions} & * & 8.33 & \multirow{2}{*}{89.17}\\ 
        & {\checkmark} &97.50 &\\
        \midrule
        \multirow{2}{0.22\linewidth}{Only} & * & 5.56 & \multirow{2}{*}{84.44}\\ 
        & {\checkmark} & 90.00 &\\
        \midrule
        \multirow{2}{0.25\linewidth}{Sentential negation} & * & 27.78 & \multirow{2}{*}{52.22}\\ 
        & {\checkmark} & 80.00 &\\
        \midrule
        \multirow{2}{0.25\linewidth}{Simple questions} & * & 33.33 & \multirow{2}{*}{62.97}\\ 
        & {\checkmark} & 96.30 &\\
        \midrule
        \multirow{2}{0.25\linewidth}{Superlatives} & * & 8.33 & \multirow{2}{*}{66.67}\\ 
        & {\checkmark} & 75.00 &\\
        \midrule
        \multirow{2}{0.25\linewidth}{Quantifiers} & * & 4.17 & \multirow{2}{*}{50.83}\\ 
        & {\checkmark} & 55.00 &\\
        \bottomrule
    \end{tabular}
    \caption{Results from MTurk validation. `Environment' is the name of the licensing environment and `label' is whether the sentence was intended as acceptable ({\checkmark}) or unacceptable (*). The results of the validation ratings is in `\% accept' and represents the majority vote for each sentence as acceptable/unacceptable and then averaged to give the percentage of times a sentence in a given condition was rated as acceptable by the MTurk raters. `Diff' is calculated from the \% of acceptable sentences rated acceptable minus the \% of unacceptable sentences rated acceptable (100 is a perfect score, 0 means there is no difference).}
    \label{tab:mturk_validation}
\end{table}

\end{document}